%% file: main.tex
\documentclass[11pt,a4paper]{article}
\usepackage[hyperref]{eacl2021}
\usepackage{times}
\usepackage{latexsym}

\usepackage{microtype}
\usepackage{booktabs} 

\usepackage{blindtext}
\usepackage{multirow}
\usepackage{graphicx}

\usepackage{amsmath}
\usepackage{amssymb}
\usepackage{mathtools}
\usepackage{breqn}
\usepackage[normalem]{ulem}

\usepackage[export]{adjustbox}
\usepackage{float}
\restylefloat{table}

\usepackage{wrapfig}

\newcommand\comment[1]{}
\newcommand\vv{\mathbf{v}}
\newcommand\ov{\mathbf{o}}
\newcommand\pv{\mathbf{p}}
\newcommand\qv{\mathbf{q}}

\aclfinalcopy 
\def\aclpaperid{1352} 

\title{BERTese: Learning to Speak to BERT}

\author{Adi Haviv$^{1}$~~~~~~~~~Jonathan Berant$^{1,2}$~~~~~~~~~Amir Globerson$^1$\\ \\
$^1$School of Computer Science, Tel Aviv University \\
$^2$Allen Institute for AI \\
\small{\texttt{ \{adi.haviv,joberant\}@cs.tau.ac.il, gamir@post.tau.ac.il}}
}

\date{}

\begin{document}
\maketitle

\input{sections/00_abstract.tex}
\input{sections/01_intro}
\input{sections/02_related_work}
\input{sections/03_bertese_model}
\input{sections/04_exeriments}
\input{sections/05_results}
\input{sections/06_conclusion}
\input{sections/07_acknowledgements}


\bibliography{anthology,eacl2021}
\bibliographystyle{acl_natbib}

\end{document}

%% file: sections/00_abstract.tex
\begin{abstract}
Large pre-trained language models have been shown to encode large amounts of world and commonsense knowledge in their parameters, leading to substantial interest in methods for extracting that knowledge. In past work, knowledge was extracted by taking manually-authored queries and gathering paraphrases for them using a separate pipeline. In this work, we propose a method for automatically rewriting queries into ``BERTese'', a paraphrase query that is directly optimized towards better knowledge extraction. To encourage meaningful rewrites, we add auxiliary loss functions that encourage the query to correspond to actual language tokens. 
We empirically show our approach outperforms competing baselines, obviating the need for complex pipelines. Moreover, BERTese provides some insight into the type of language that helps language models perform knowledge extraction.
\end{abstract}

%% file: sections/01_intro.tex
\section{Introduction}
\label{sec:intro}

\begin{figure}[t]
    \centering
    \includegraphics[width=.45\textwidth,left]{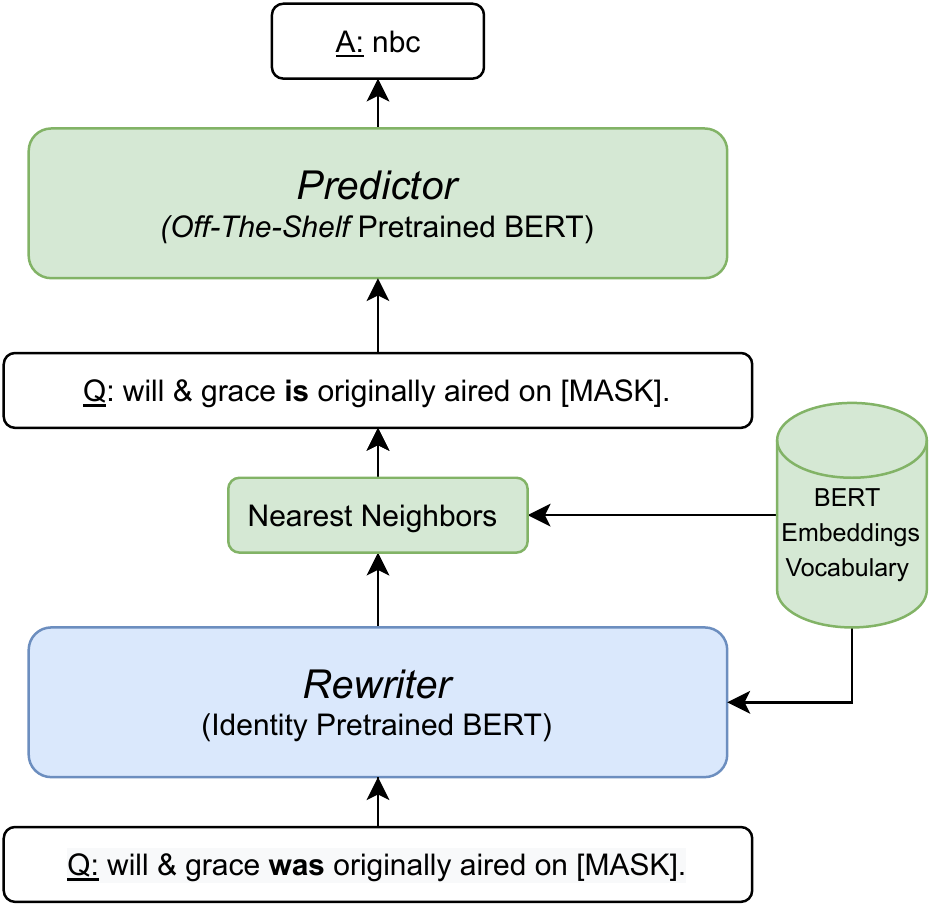}
    \caption{The BERTese Model. The model takes an input query, rewrites it, and feeds the output to a pre-trained BERT model. The untrained components are marked in green, and the blue component is trained.}
    \label{fig:bertese_fig}
\end{figure}

Recent work has shown that large pre-trained language models (LM), trained with a masked language modeling (MLM) objective \cite{Devlin2019BERTPO, Liu2019RoBERTaAR, Lan2020ALBERTAL,Sanh2019DistilBERTAD,Conneau2020UnsupervisedCR}, encode substantial amounts of world knowledge in their parameters. This has led to ample research on developing methods for extracting that knowledge \cite{Petroni2019LanguageMA, Petroni2020HowCA, Jiang2019HowCW, Bouraoui2020InducingRK}. The most straightforward approach is to present the model with a manually-crafted query such as \emph{``Dante was born in [MASK]''} and check if the model predicts \emph{``Florence''} in the \emph{[MASK]} position. However, when this fails, it is difficult to determine if the knowledge is absent from the LM or if the model failed to understand the query itself. For example, the model might return the correct answer if the query is \emph{``Dante was born in the city of [MASK]''}. 

Motivated by the above observation, we ask: can we automatically find the best way to ``ask'' an LM about its knowledge? We refer to this challenge as speaking ``BERTese''. In particular, we ask how to rewrite a knowledge-seeking query into one that MLMs understand better, where understanding is manifested by providing a correct answer to the query. 

Prior work \cite{Jiang2019HowCW} tackled this problem using a 2-step pipeline, where first a small list of paraphrase templates is collected using external resources, and then a model learns to extract knowledge by aggregating information from paraphrases of the input query. In this work, we propose a more general approach, where the model learns to rewrite queries, directly driven by the objective of knowledge-extraction.

Figure~\ref{fig:bertese_fig} provides an overview of our approach. Our model contains a BERT-based \emph{rewriter}, which takes a query as input, and outputs for each input position a new token, which is its rewrite. This new query is fed into a different BERT \emph{predictor} from which the answer is extracted.  Importantly, the downstream predictor BERT is a fixed pre-trained model, and thus the goal is to train the rewriter to produce queries for which the predictor outputs the correct answer.

A technical challenge is that outputting discrete tokens leads to a non-differentiable model, which we tackle by adding a loss term that encourages the rewriter's output to be similar to BERT token embeddings. Moreover, we must guarantee that the BERTese query contains the \emph{[MASK]} token from which the answer will be read. To achieve this, we first add an auxiliary loss term that encourages the model to output precisely one masked token in the query rewrite. We then add a layer that finds 
the token index that most closely resembles \emph{[MASK]}, and this is where we expect the correct answer to be completed. Training of this selection process is done using the straight-through estimator \cite{Hinton2012,Bengio2013EstimatingOP}.

We evaluate our approach on the LAMA dataset \cite{Petroni2019LanguageMA}, and show that our model significantly improves the accuracy of knowledge extraction. Furthermore, many of the rewrites correspond to consistent changes in query wording (e.g., changing tense), and thus provide information on the types of changes that are useful for extracting knowledge from BERT. 
While we experiment on BERT, our method is generic and can be applied to any MLM.

Taken together, our results demonstrate the potential of rewriting inputs to language models for both obtaining better predictions, and for potentially gaining insights into how knowledge is represented in these models. Our code can be downloaded from \url{https://github.com/adihaviv/bertese}.

%% file: sections/02_related_work.tex
\section{Related Work}
\label{sec:related_work}

Choosing the right language for extracting world knowledge from LMs has attracted much interest recently. First, \newcite{Petroni2019LanguageMA} observed that MLMs can complete simple queries with correct factual information. \newcite{Jiang2019HowCW} and \newcite{Heinzerling2020LanguageMA} then showed that in the zero-shot setting, small variations to such queries can lead to a drop in fact recall.  
Orthogonally, another line of research focused on query reformulation for standard Question Answering (QA) tasks. \newcite{Gan2019ImprovingTR} demonstrated that even minor query modifications can lead to a significant decrease in performance for multiple QA models and tasks. \newcite{Buck2018AskTR} showed that it is possible to train a neural network to reformulate a question using Reinforcement Learning (RL), optimizing the accuracy of a black-box QA system. Similarly, \newcite{Nogueira2017TaskOrientedQR} used RL to create a query reformulation system that maximizes the recall of a black-box information retrieval engine. 

\newcite{Jiang2019HowCW} proposed an ensemble method for query reformulation from LMs, that includes: (1) mining new queries, (2) using an off-the-shelf pre-trained translation model to collect additional paraphrased queries with back-translation, and (3) using a re-ranker to select one or more of the new queries. They then feed those queries to BERT to get the masked token prediction. 

In this work, we take the idea of \citet{Jiang2019HowCW} a step forward and train a model in an end-to-end fashion to generate rephrased queries which are optimized to maximize knowledge extraction from the MLM.\footnote{Although knowledge retrieval has been investigated in autoregressive models as well, similar to \citet{Jiang2019HowCW}, in this work we focus on MLMs only, as AR-LM only predict an answer if the masked token is at the end of the query.}
 Independent of this work, \newcite{shin-etal-2020-autoprompt} recently proposed to automatically generate prompts for extracting knowledge from a pre-trained language model using a gradient-based search method, illustrating gains on multiple tasks.

%% file: sections/03_bertese_model.tex
\section{The BERTese Model}
\label{sec:bertese}
Recall that our goal is to build a model that takes as input a query in natural language, and re-writes it into a query that will be fed as input to an existing BERT model. 

We refer to the above re-writing model as the \textit{rewriter} and the existing BERT model as the \textit{predictor}. We note that both input and output queries should include the token [MASK]. For example the input could be \emph{``Obama was born in [MASK]''} and the output \emph{``Obama was born in the state of [MASK]''}.

We first describe the behaviour of our model at inference time (see Figure~\ref{fig:bertese_fig}).
Given a query, which is a sequence of tokens, $S = (s_1, \dots, s_n)$, we map $S$ into a sequence of vectors $Q(S) \in \mathbb{R}^{d \times n}$ using BERT's embeddings of dimensionality $d$. This input is fed into a (BERT-based) stack of transformer layers that outputs a new sequence of vectors $\hat{Q}(S) \in \mathbb{R}^{d \times n}$. 

To obtain vectors that can be used as input to the predictor, we need to map the vectors in each position to their nearest neighbor in the set of BERT embeddings. Specifically, let $B_V$ be the set of BERT embeddings, and let $\hat{Q}_i \in \mathbb{R}^d$ be the re-written vector in position $i$. We map $\hat{Q}_i \in \mathbb{R}^d$ to $\arg\min_{\vv \in B_V}\left( \left\|v-\hat{Q}_i \right\|_2^2 \right)$. We next pass the re-written query into the pre-trained predictor BERT model, and obtain an answer from the most probable token in the masked position.

Training this model involves two technical challenges. First, the nearest-neighbor operation is non-differentiable. Second, to obtain the prediction of the \verb|[MASK]| token, we need to guarantee that the rewriter generates a \verb|[MASK]| token, and know its position (because this is where the ground-truth answer should be predicted). We overcome these by adding two auxiliary loss functions. The first encourages the model to output vectors that are similar to BERT embeddings (thus reducing the loss in the nearest neighbor operation), and the second encourages the model to output one masked token.

Finally, we apply the straight-through estimator, which allows us to feed discrete word representations into the predictor and backpropagate the signal back to the rewriter. We next provide more details on the terms in our loss function used to train the rewriter.

\paragraph{Valid Token Loss:} At training time we do not apply the non-differentiable nearest-neighbor operation. Thus, we would like the vectors $\hat{Q}(S)$ output by the rewriter to be as close as possible to valid BERT embeddings. This loss is the average over tokens of the distance between a re-written query token and its nearest neighbor:
\begin{equation}
f_1(S) = \frac{1}{|\hat{Q}(S)|} \sum_{\qv \in \hat{Q}(S)}\min_{\vv \in B_V}\left( \left\|\vv-\qv \right\|_2^2 \right).
\label{eq:training-objectives_f_1}
\end{equation}

\paragraph{Single [MASK] Loss:} The output of the rewriter must contain the \emph{[MASK]} token, so that the predictor can extract an answer from this token. 
To encourage the rewriter to output a [MASK] we add a loss as follows. We define the following ``softmin'' distribution over $i\in\{1,\ldots,|\hat{Q}(S)|\}$:
\begin{equation}
m_i(S) = \frac{e^{-\beta \left\|B_{[MASK]}-\hat{Q}_i(S) \right\|_2^2}}{\sum_j e^{-\beta \left\|B_{[MASK]}-\hat{Q}_j(S) \right\|_2^2}},
\end{equation}
where $\beta$ is a trained parameter.
The maximum value of this distribution will be highest when there is a single index $i$ that is closest to the embedding of [MASK] (if there are two maxima, they will both have equal values). 
Thus the loss we consider is:
\begin{equation}
f_2(S) = - \max_i m_i(S).
\label{eq:training-objectives_f_2}
\end{equation}

\paragraph{Prediction Loss:} The predictor should return the gold answer $y$ when given $\hat{Q}$ as input. Without non-differentiability, we could find the index of the [MASK] token in $\hat{Q}$, and use cross-entropy loss between the output distribution of the predictor in that index and the gold answer $y$. To remedy this, we use a differentiable formulation, combined with the straight-through estimator (STE) \cite{Bengio2013EstimatingOP}:
Let $\ov_i$ be the output distribution at the $i^{\text{th}}$ position of the predictor, and let $\ell(y,\pv)$ be the cross-entropy between the one-hot distribution corresponding to $y$ and a distribution $\pv$. Then, we use the loss:
\begin{equation}
    f_{CE}(S,y) = \sum_{i} m_i(S)\ell(y, \ov_i).
    \label{eq:label-loss}
\end{equation}
Thus, if $m$ is a one-hot on the index corresponding to [MASK], the loss will be the desired cross-entropy between the gold answer and the predicted distribution. We optimize this objective using the STE. Namely, in the forward pass, we convert $m$ to a one-hot vector.

Our final training loss is the sum of the above three loss terms:
\begin{equation}
L(S,y) = f_{CE}(S,y) + \lambda_1 \cdot f_1(S) + \lambda_2 \cdot f_2(S).
\label{eq:training-objectives}
\end{equation}
The weights $\lambda_1, \lambda_2$ are tuned using cross-validation.

To summarize, the main challenge is that the rewriter output needs to be optimized to predict the correct label for the [MASK] token (Eq. \ref{eq:label-loss}). However, the [MASK] token needs to appear once in the rewriter output. In order to enforce the above, the ``Single [MASK] Loss'' (Eq.~\ref{eq:training-objectives_f_2}) is used. In addition, in order for the rewriter output to be a valid sentence, the ``Valid Token Loss'' (Eq.~\ref{eq:training-objectives_f_1}) is added. This encourages the model to output tokens that are close to BERT input embeddings. This is done by minimizing the distance between each rewriter vector to some vector in the BERT input embedding dictionary.

\paragraph{Rewriter pre-training}
We initialize the rewriter with a BERT-based model, additionally fine-tuned to output the exact word embeddings it received as input (i.e., fine-tuned to the identity mapping). Thus, when training for knowledge extraction, the rewriter is initialized to output exactly the query it received as input.

%% file: sections/04_exeriments.tex
\section{Experiments}
\label{sec:experiments}

\begin{table*}[t]
\centering
\begin{tabular}{lllll}
\toprule
\textbf{Corpus}    & \textbf{BERT} & \textbf{FT-BERT} & \textbf{LPAQA} & \textbf{BERTese} \\ \midrule
\textbf{T-REx}     & 31.1          & 36               & 34.1           & \textbf{38.3}    \\
\bottomrule
\end{tabular}
\caption{Mean precision at one (P@1) for three baselines and our BERTese model on the T-REx dataset.}
\label{tb:results}
\end{table*}

\begin{table*}[h]
\small
\centering
\begin{tabular}{@{}lll@{}}
\toprule
\textbf{Modification~} & \textbf{Original Masked Query} & \textbf{Bertese Masked Query}    \\
\midrule
\vspace{0.1cm}
\textit{"!" removed} & {yahoo\textbf{!} tech is owned by [MASK].}         & {yahoo tech is owned by [MASK].}   \\\vspace{0.1cm}
\textit{verb patterns} & {{\textbf{working~} dog is a subclass of [MASK].}} & {{\textbf{work~} dog is a subclass of [MASK].}}  \\\vspace{0.1cm}
\textit{was} $\rightarrow$ \textit{is}  &  {will \& grace \textbf{was }originally aired on [MASK].} &  {will \& grace \textbf{is} originally aired on [MASK].}    \\\vspace{0.1cm}

\textit{a} $\rightarrow$ \textit{the}  &  
{tom terriss is \textbf{a} [MASK] by profession.} &  {tom terriss is \textbf{the} [MASK] by profession.}    \\\vspace{0.1cm}

\textit{rephrasing~} & istanbul hezarfen~\textbf{airfield~}is named after [MASK]. & istanbul hezarfen~\textbf{airport~}is named after [MASK]. \\\vspace{0.1cm}

\textit{token} $\rightarrow$ \textit{[SEP]}  & \small{\textbf{lub}ka kolessa plays [MASK].} & \small{\textbf{[SEP]}ka kolessa plays [MASK].} \\
\bottomrule 
\end{tabular}
\caption{Examples of rewrites from the T-REx test-set, where the original query resulted in a wrong answer, and the BERTese rewrite resulted in correct one.}
\label{tb:examples}
\end{table*}

\paragraph{Experimental setup}
We conduct our experiments on the LAMA dataset \cite{Petroni2019LanguageMA,Jiang2019HowCW}, a recently introduced unsupervised knowledge-extraction benchmark for pre-trained LMs. 
LAMA is composed of a collection of cloze-style queries about relational facts with a single token answer. As in \citet{Jiang2019HowCW}, we limit our main experiment to the T-REx \cite{ElSahar2018TRExAL} subset. The T-REx dataset is constructed out of 41 relations, each associated with at most 1000 queries, all extracted from Wikidata.

For training our model, we use a separate training set, created by \newcite{Jiang2019HowCW}, called T-REx-train. This dataset is constructed from Wikidata and has no overlap with the original T-REx dataset. We evaluate our model on the complete T-REx dataset. 

\paragraph{Implementation Details}
Both the rewriter and the predictor are based on BERT$_{\text{base}}$ with the default settings from the Huggingface \cite{Wolf2019HuggingFacesTS} platform.
We optimize BERTese using AdamW with an initial learning rate of 1e-5. We train the model on a single 32GB NVIDIA V100 for 5 epochs with a batch size of 64. 
For the loss coefficients (see Eq. {\eqref{eq:training-objectives})} we set $\lambda_1=0.3$ and  $\lambda_2=0.5$. 

\paragraph{Baselines}
We compare our method to three baselines: (a) BERT - 
A BERT$_{\text{base}}$ model without any fine-tuning, as evaluated in \newcite{Petroni2019LanguageMA}. (b) LPAQA - The model proposed by \newcite{Jiang2019HowCW}, based on mining additional paraphrase queries. We report results on a single paraphrase.\footnote{It is possible to improve results by aggregating over multiple rewrites, but our focus is on a single rewrite.} (c) FT-BERT: An end-to-end differentiable BERT$_{\text{base}}$ model, explicitly fine-tuned on T-REx-train to output the correct answer. This model, like ours, is trained for knowledge extraction, but does this internally, without exposing an interpretable intermediate textual rewrite. 

%% file: sections/05_results.tex
\paragraph{Results}
\label{sec:results}

We use the same evaluation metrics as \newcite{Petroni2019LanguageMA}  and report precision at one (P@1) macro-averaged over relations (we first average within relations and then across relations). As shown in Table \ref{tb:results}, BERTese outperforms all three baselines. Compared to the zero-shot setting, where BERT is untrained on any additional data, we improve performance from $31.1 \rightarrow 38.3$. Our model also outperforms a BERT model fine-tuned for knowledge extraction on the same data as our model ($36 \rightarrow 38.3$). Last, we outperform the BERT$_{\text{base}}$ version of LPAQA by more than 4 points.

Table~\ref{tb:examples} presents example rewrites that are output by our model. It can be seen that rewrites are usually semantically plausible, and make small changes that are not meaningful to humans, but seem to help extract information from BERT, such as $\emph{was} \rightarrow \emph{is}$ and $\emph{a} \rightarrow \emph{the}$. In some cases, rewrites can be interpreted, for example, replacing the word \emph{airfield} with the more frequent word \emph{airport}.

\paragraph{Ablation Study}
 In Table \ref{tb:ablation_results} we present P@1 results on the T-REx test set after ablating different parts of the loss function. We keep the same label loss, same rewriter pretraining scheme, hyperparameters, and inference process. We show that removing all auxiliary losses hurts performance significantly on the T-REx dataset. Next, we evaluate the impact of removing the ``Single [MASK] Loss'', and report a drop from $38.3$ to $37.3$. In addition, when further observing the rewrites the model produces, we find that those will have in some cases more than one [MASK] token. Overall, the results show that having just one of the loss terms substantially improves the performance (either ``Valid Token Loss'' or ``Single [MASK] Loss''), but using both losses further improves accuracy.

\begin{table}[H]
\centering
\begin{tabular}{lc}
\toprule
\textbf{Ablation}   & \textbf{P@1} \\ \midrule
No auxilary losses & 25.3 \\
SML     & 36.6  \\
VTL     & 37.5  \\
SML + VTL (BERTese)    & 38.3  \\

\bottomrule
\end{tabular}
\caption{Ablation experiments on T-REx. We abbreviate  the "Single [MASK] token" as SML and the "Valid Token Loss" as VTL.}
\label{tb:ablation_results}
\end{table}

\paragraph{Part Of Speech Analysis} To better understand what types of changes our rewriter performs, Table \ref{tb:pos_analysis} shows the distribution over part-of-speech-tags replaced by the rewriter. We show all part-of-speech tags for which the frequency is higher than 1\%. More than 70\% of the replacements are nouns and verbs, which carry substantial semantic content. Interestingly, 15\% of the replacements are determiners, which bear little semantic content.

\begin{table}[H]
\centering
\begin{tabular}{ll}
\toprule
\textbf{POS Tag} & \textbf{Frequency} \\ \midrule
NN   & 47.6\%   \\
VBN   & 23\%    \\
DT   & 15.3\%    \\
JJ   & 4.4\%    \\
CD   & 3\%    \\
NNP   & 1.7\%   \\
NNS  & 1.3\%   \\

\bottomrule
\end{tabular}
\caption{Part-of-speech analysis of rewrites from the T-REx test-set.}
\label{tb:pos_analysis}
\end{table}

%% file: sections/06_conclusion.tex
\section{Conclusion}
We presented an approach for modifying the input to a BERT model, such that factual information can be more accurately extracted. Our approach uses a trained rewrite model that is optimized to maximize the accuracy of its rewrites, when used as input to BERT. Our rewriting scheme indeed turns out to produce more accurate results than baselines. Interestingly, our rewrites are fairly small modifications, highlighting the fact that BERT models are not invariant to these edits.

Our approach is not limited to knowledge extraction. It can, in principle, be applied to BERT in general question answering datasets and even language modeling. In the former, we can change the predictor to a multiple-choice QA pretrained BERT and exclude the single [MASK] token loss. In the latter, we can for example envision a case where rewriting a sentence can make it easier to complete a masked word.

Our empirical setting focuses on the LAMA dataset, where a single mask token prediction is required. There are several possible extensions to multiple masks, and we leave these for future work.
Finally, it will be interesting to test the approach on other masked language models such as RoBERTa \cite{Liu2019RoBERTaAR} and ERNIE \cite{Zhang2019ERNIEEL}, a MLM that is enhanced with external entity representations.

%% file: sections/07_acknowledgements.tex
\section{Acknowledgments}
We thank Yuval Kirstain for his contribution in both the Rewriter Pretraining and overall valuable comments, and Edo Cohen-Karlik for his insightful suggestions. This project received funding from the European Research Council (ERC) under the European
Union Horizon 2020 research and innovation programme (grant ERC HOLI 819080).